\documentclass[letterpaper, 10 pt, conference]{ieeeconf} 
\IEEEoverridecommandlockouts
\usepackage{graphics}
\usepackage{amsmath}
\usepackage{amssymb}
\usepackage{hyperref}
\usepackage{wrapfig}
\usepackage{balance}
\usepackage[font=small]{caption}
\usepackage{multirow}
\usepackage{wrapfig}
\usepackage{todonotes}

\newcommand\defeq{\mathrel{\stackrel{\makebox[0pt]{\mbox{\normalfont\scriptsize def}}}{:=}}}

\usepackage{tikz}
\newcommand\copyrighttext{%
  \footnotesize © 2025 IEEE.  Personal use of this material is permitted.  Permission from IEEE must be obtained for all other uses, in any current or future media, including reprinting/republishing this material for advertising or promotional purposes, creating new collective works, for resale or redistribution to servers or lists, or reuse of any copyrighted component of this work in other works.}
\newcommand\copyrightnotice{%
\begin{tikzpicture}[remember picture,overlay]
\node[anchor=south,yshift=10pt] at (current page.south) {\fbox{\parbox{\dimexpr\textwidth-\fboxsep-\fboxrule\relax}{\copyrighttext}}};
\end{tikzpicture}%
}

\title{\LARGE \bf
FuzzRisk: Online Collision Risk Estimation for Autonomous Vehicles based on Depth-Aware Object Detection via Fuzzy Inference
}

\author{Brian Hsuan-Cheng Liao$^{\dagger\ddagger}$, Yingjie Xu$^{\ddagger}$, Chih-Hong Cheng$^{\mathsection\sharp}$, Hasan Esen$^{\dagger}$, Alois Knoll$^{\ddagger}$ % 
\thanks{$^{\dagger}$DENSO AUTOMOTIVE Deutschland GmbH, Germany} %
\thanks{$^{\ddagger}$Technical University of Munich, Germany} %
\thanks{$^{\mathsection}$Chalmers University of Technology, Sweden} %
\thanks{$^{\sharp}$University of Gothenburg, Sweden} %
\thanks{Correspondence to \tt\small {h.liao}@eu.denso.com}
}

\begin{document}

\maketitle
\copyrightnotice

\begin{abstract}

This paper presents a novel monitoring framework that infers the level of collision risk for autonomous vehicles (AVs) based on their object detection performance. The framework takes two sets of predictions from different algorithms and associates their inconsistencies with the collision risk via fuzzy inference. The first set of predictions is obtained by retrieving safety-critical 2.5D objects from a depth map, and the second set comes from the ordinary AV's 3D object detector. We experimentally validate that, based on Intersection-over-Union (IoU) and a depth discrepancy measure, the inconsistencies between the two sets of predictions strongly correlate to the error of the 3D object detector against ground truths. This correlation allows us to construct a fuzzy inference system and map the inconsistency measures to an AV collision risk indicator. In particular, we optimize the fuzzy inference system towards an existing offline metric that matches AV collision rates well. Lastly, we validate our monitor's capability to produce relevant risk estimates with the large-scale nuScenes dataset and demonstrate that it can safeguard an AV in closed-loop simulations.

\end{abstract}

\section{Introduction}
\label{sec:intro}

Over the past decade, autonomous vehicles (AVs) have attained great development and can be seen on public roads nowadays. However, it is still possible to hear AV accidents, especially in corner cases such as severe weather conditions or the emergence of rare objects~\cite{vanzura2024autonomous}. To ensure the safety of AVs and allow their wider deployment, it is important to have run-time monitors that can identify such performance-hindering situations. Correspondingly, regulations and industrial standards such as EU AI Act~\cite{aiact} and ISO~21448~\cite{iso21448} also demand the inclusion of monitoring mechanisms in safety-critical autonomous systems.

In the literature, in fact, one can find various run-time AV monitoring techniques.
Focusing on planning and control, many works draw collision risks based on ego and traffic information, e.g., deviation of the driving path or time to collision towards other agents~\cite{chia2022risk}. 
However, these approaches assume perfect perception, which is usually not the case. 
Our work attempts to relax the assumption and identify hazards as early as possible at the perception phase of the AV software stack.
Several studies have done so similarly, monitoring the object detection function, given its pivotal role for an AV to operate.
Their algorithms, for example, check the spatial or temporal consistency of the set of detected objects~\cite{chen2021monitoring,balakrishnan2021percemon}.
While their results appear effective, one crucial limitation is the relevance between the identified anomalies and the actual safety risk of the AV.
For instance, these object detection monitors may raise a warning for an object that is actually far from the AV and less relevant.
Likewise, they only examine the set of detected objects and ignore potential misses (i.e., false negatives of the object detector), which possibly pose greater risks to safety.

\begin{figure}[]
    \centering
    \includegraphics[width=0.98\linewidth]{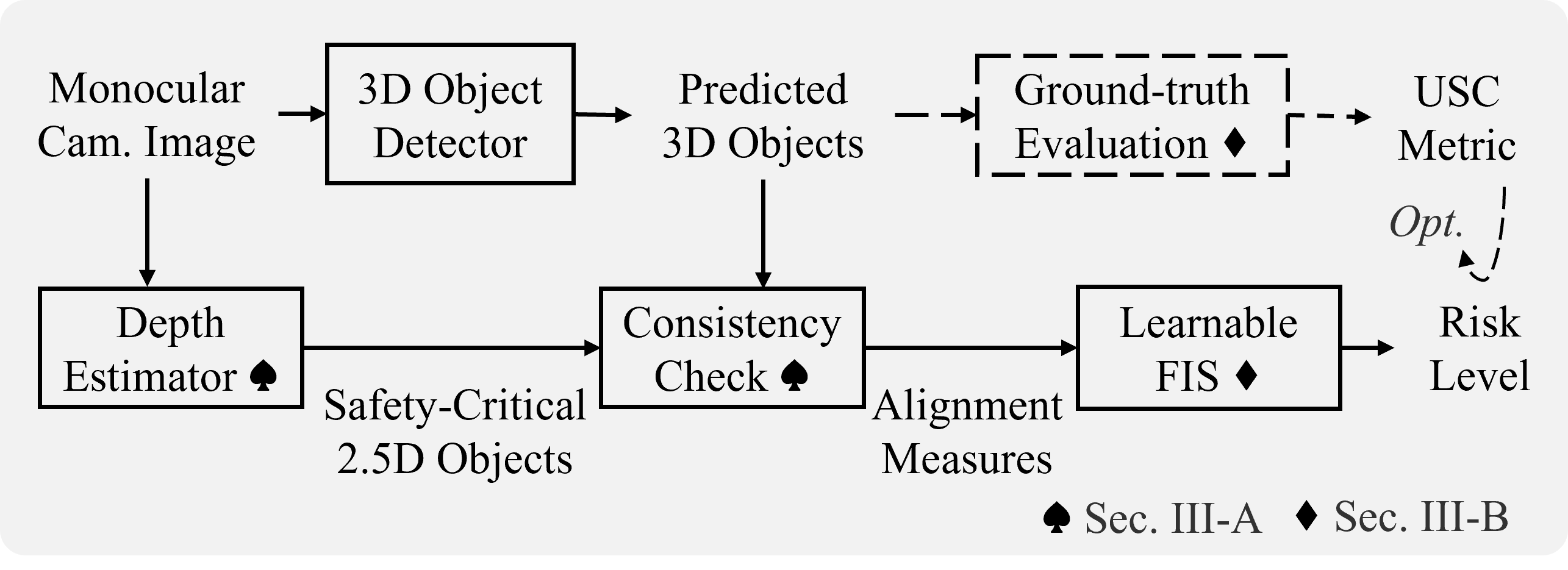}
    \caption{Our overall monitoring framework uses alignment measures between two sets of predictions to infer ego vehicle's collision risks. The dashed lines mark an offline optimization process using a previously presented risk-correlating metric, USC~\cite{liao2024usc}.}
    \label{fig:framework}
\end{figure}

Hence, in this work, we aim to find safety-relevant errors of the object detector and use them to characterize an AV's collision risks. 
Fig.~\ref{fig:framework} depicts the overall framework.
In particular, our previous work presented a design-time safety-oriented metric, uncompromising spatial constraints (USC), and showed that it matches AV collision rates well using simulations~\cite{liao2024usc}. 
This work then utilizes USC during offline optimization to eventually obtain a reliable collision risk indicator that can help protect the AV in operation.
To achieve this, there are two challenges:
\begin{itemize}
    \item The first and main challenge lies in the \emph{lack of ground-truth labels at run time}. 
    To overcome it, we ask the critical question of whether employing a separate object retrieval pipeline and measuring the inconsistencies from the original object detector's predictions can serve as a proxy to the ground truths.
    Considering cost factors and the recent breakthrough in monocular depth estimation, we employ the state-of-the-art ZoeDepth~\cite{bhat2023zoedepth} and implement image processing techniques to retrieve safety-critical objects.
    Our key insight, thereby, is an experimental validation that confirms the inconsistencies between the two sets of predictions, measured by Intersection-over-Union (IoU) and depth discrepancy, are \emph{closely linked} to the errors of the 3D object detector against ground truths.
    \item With the confirmed correlation, the second challenge is \emph{how to associate them with the desired risk quantifier}. Our solution is to use fuzzy logic, which can flexibly model non-linear functions while tolerating potential imprecision~\cite{zadeh1988fuzzy}. More concretely, we build a fuzzy inference system (FIS) using three distinct knowledge- and data-driven approaches and finally find the \emph{global particle swarm optimization} algorithm gives the best results.
\end{itemize}

To demonstrate the efficacy of our framework, we run it with a state-of-the-art 3D object detector, PGD~\cite{wang2021probabilistic}, across the large-scale nuScenes dataset~\cite{caesar2020nuscenes} in six conditions, including nominal scenes, night scenes, scenes with rare objects, and scenes augmented with rain, snow, or Gaussian noises. In addition, we implement our monitor on a baseline AV in simulation and showcase that it can indeed infer a useful risk indicator that helps protect the AV.
Altogether, our work can be seen as a novel attempt to derive safety-relevant collision risk estimates from the object detection function.
Moreover, as we shall show in later sections, the framework offers good interpretability and adaptability, allowing developers to continuously analyze and improve the monitor.

\section{Related Work}
\label{sec:related_work}

Monitoring of AV functions and operations has been a long-standing research field~\cite{chia2022risk}. Early work focused on estimating collision risks from system dynamics, with some further demonstrating the estimated risk information can be used to adapt the behavior of the AV~\cite{zheng2018novel,khonji2020risk}. Recently, thanks to the advent of deep learning and computer vision techniques, there are results predicting collision risks in driving scenes based on camera images directly~\cite{wang2017collision,feth2018dynamic}. However, these approaches do not consider the performance of an underlying perception function. In this regard, Grimmett et al. proposed the concept of introspective perception~\cite{grimmett2016introspective}, which broadly encompasses the fields of uncertainty quantification, out-of-distribution detection, as well as performance monitoring. We briefly overview these fields in the following.

Uncertainty quantification aims to estimate a model's confidence level in its predictions. Beyond classical methods such as the softmax output~\cite{hendrycks17baseline}, Monte Carlo dropout~\cite{gal2016dropout}, or ensemble models~\cite{lakshminarayanan2017simple}, recent studies attempt to formulate uncertainty-aware loss functions and directly estimate the variance of the object detector's predictions~\cite{le2018uncertainty,nallapareddy2023evcenternet}. Still, the main challenge in uncertainty quantification is its correlation with the actual model errors or system risks during operation. While research continues to investigate related uncertainty calibration techniques, we take an analogous yet distinct approach to infer system risks from model errors and calibrate the output via an adaptable FIS.

Out-of-distribution (OoD) detection studies seek solutions other than model confidence estimation to identify anomalous data. For example, Du \textit{et al.} models class-specific features with Gaussian distributions and generates samples to train an auxiliary OoD detector~\cite{du2022vos}. Recently, Wu \textit{et al.} find it more effective to store features of training data into hyper-rectangles and use them as permissible ranges to monitor inputs~\cite{wu2024bam}.
However, it is hard to derive semantic information such as system risks from these feature-based approaches. Our usage of a depth estimator can naturally signal the monitor about such risks based on distance information.

Accordingly, performance monitors that operate on the object detector's outputs lie the closest to our work. Existing approaches can be categorized into two groups. The first group makes use of learning-based models. As an example, Rahman et al.~\cite{rahman2021online} train a neural network (NN) aside from the original object detector to estimate the typical mean Average Precision (mAP) metric. Although they achieve great regression results, mAP might not be the suitable metric for safety or collision risk indication at each frame, as pointed out by a later study~\cite{otani2022optimal}. In addition, there are still concerns about NN robustness and interpretability. Our proposal to regress a risk-relevant metric with fuzzy logic helps alleviate these concerns.
Lastly, the second group of monitoring approaches uses formal methods and specifies rules such as the consistency of predicted class labels or object locations across image frames~\cite{chen2021monitoring,balakrishnan2021percemon}. Their main constraint, as introduced, is the scope focusing on the set of predictions from an object detector only. This loses the possibility to check for detection misses. By contrast, our framework handles two sets of predictions, with one focusing on the safety-critical objects to check on the other.

Finally, our approach is related to data fusion using multiple sensing/perception channels. Readers may refer to a recent survey for a more comprehensive understanding~\cite{pandharipande2023sensing}. Compared to the mass literature, the crux of our work is the orientation towards risk indication with perception processes using only camera images.

\section{The Risk Estimation Framework}
\label{sec:method}

\begin{figure*}
    \centering
    \includegraphics[width=0.245\linewidth]{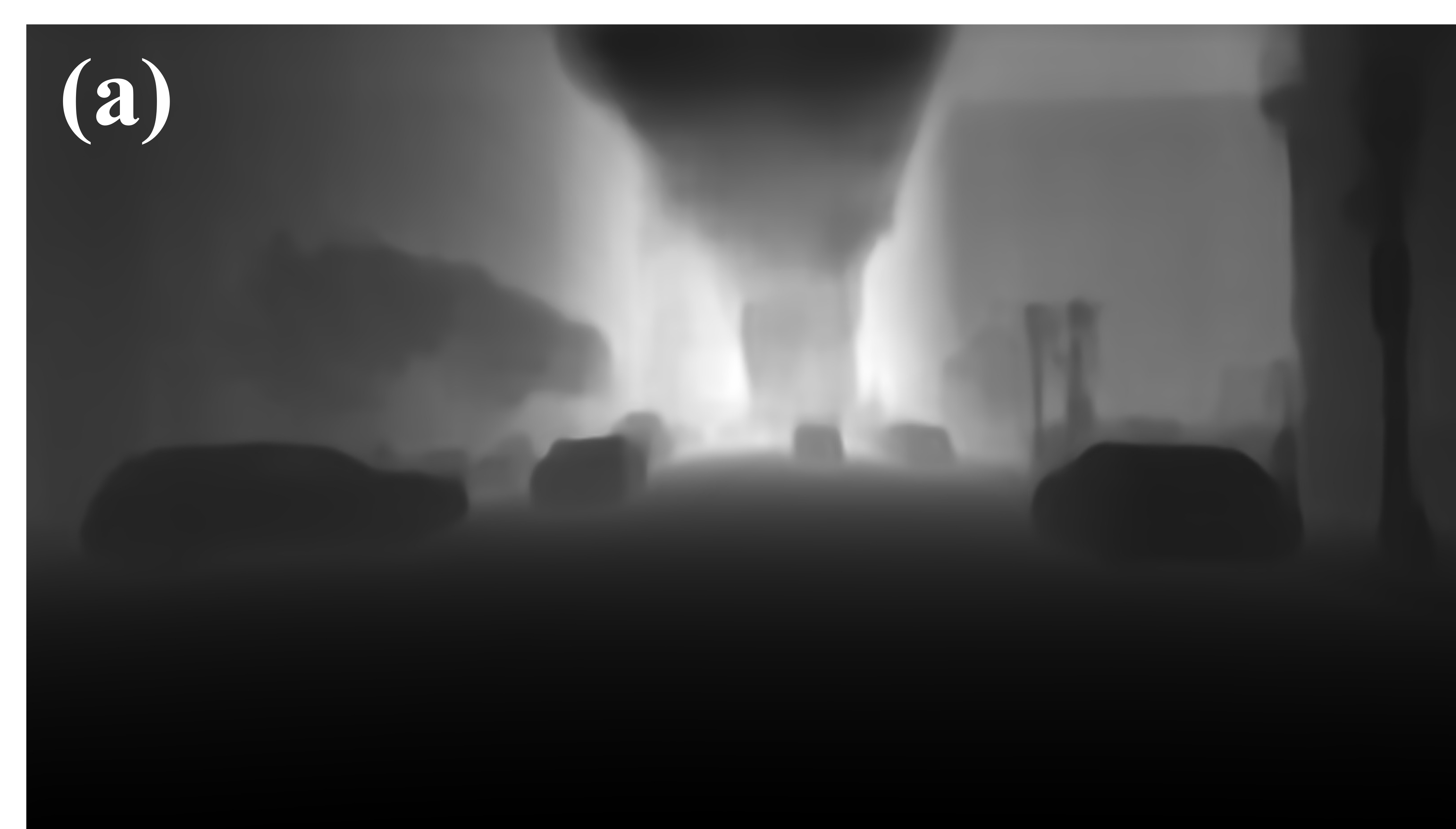}
    \includegraphics[width=0.245\linewidth]{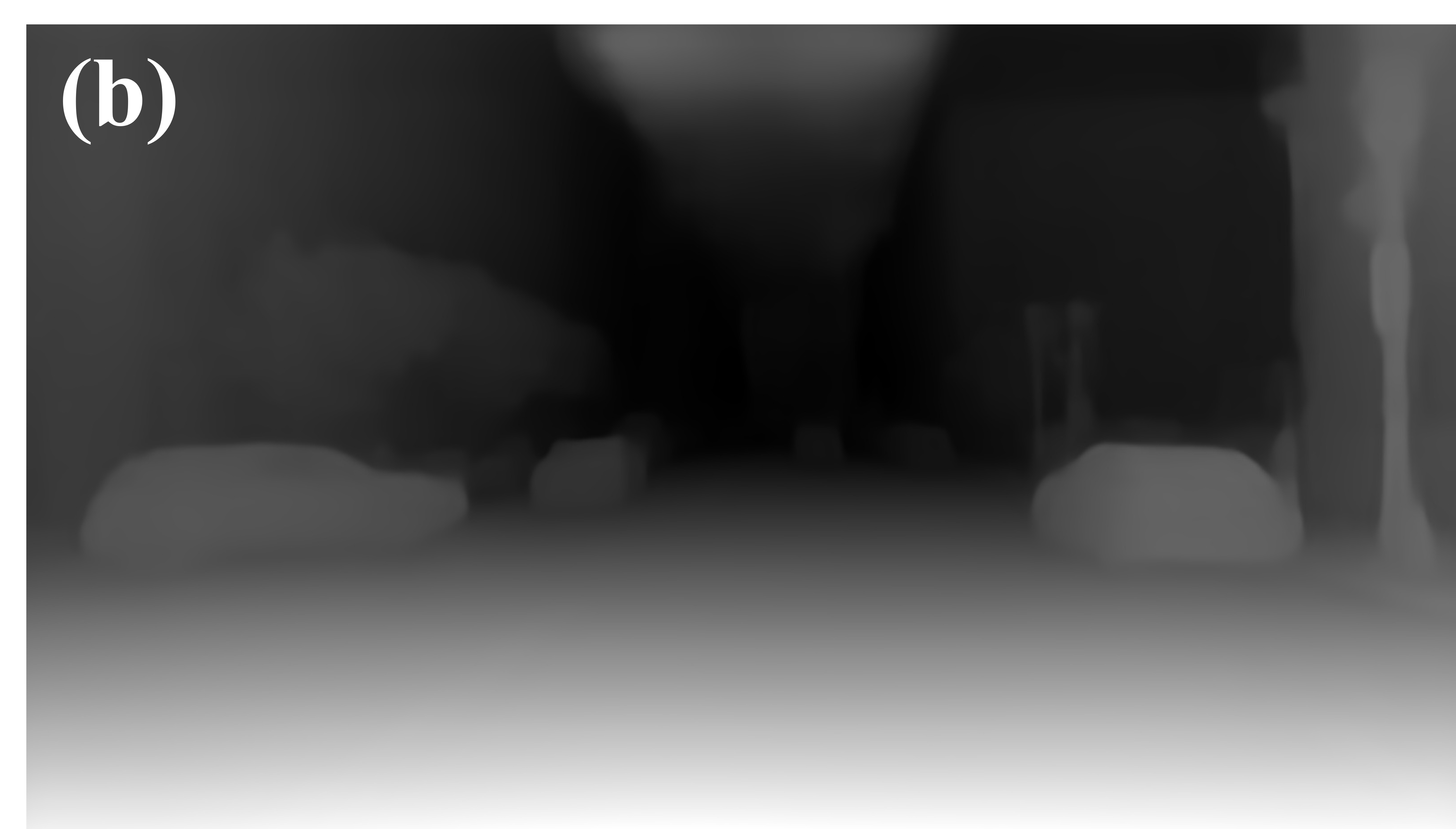}
    \includegraphics[width=0.245\linewidth]{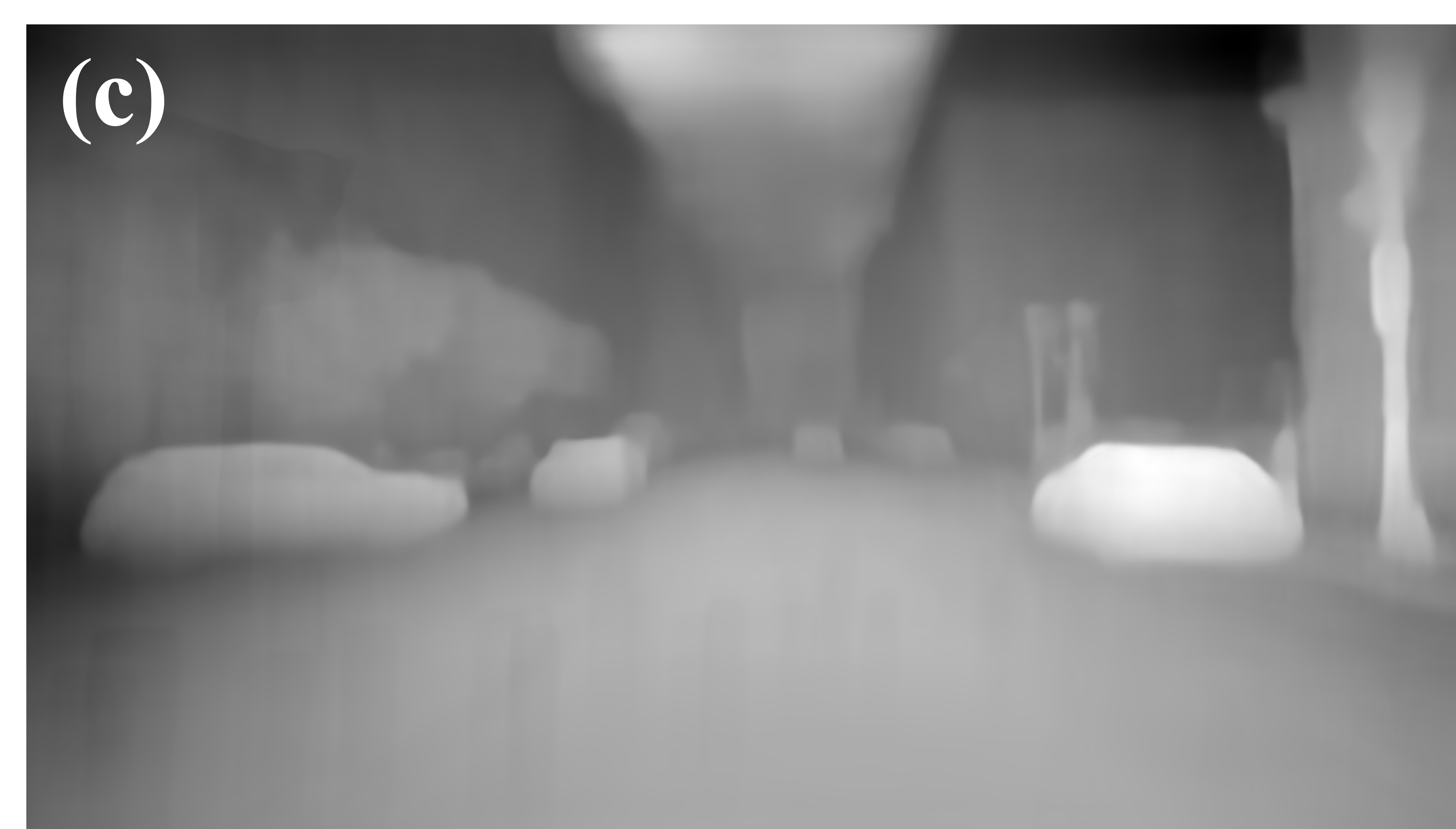}
    \includegraphics[width=0.245\linewidth]{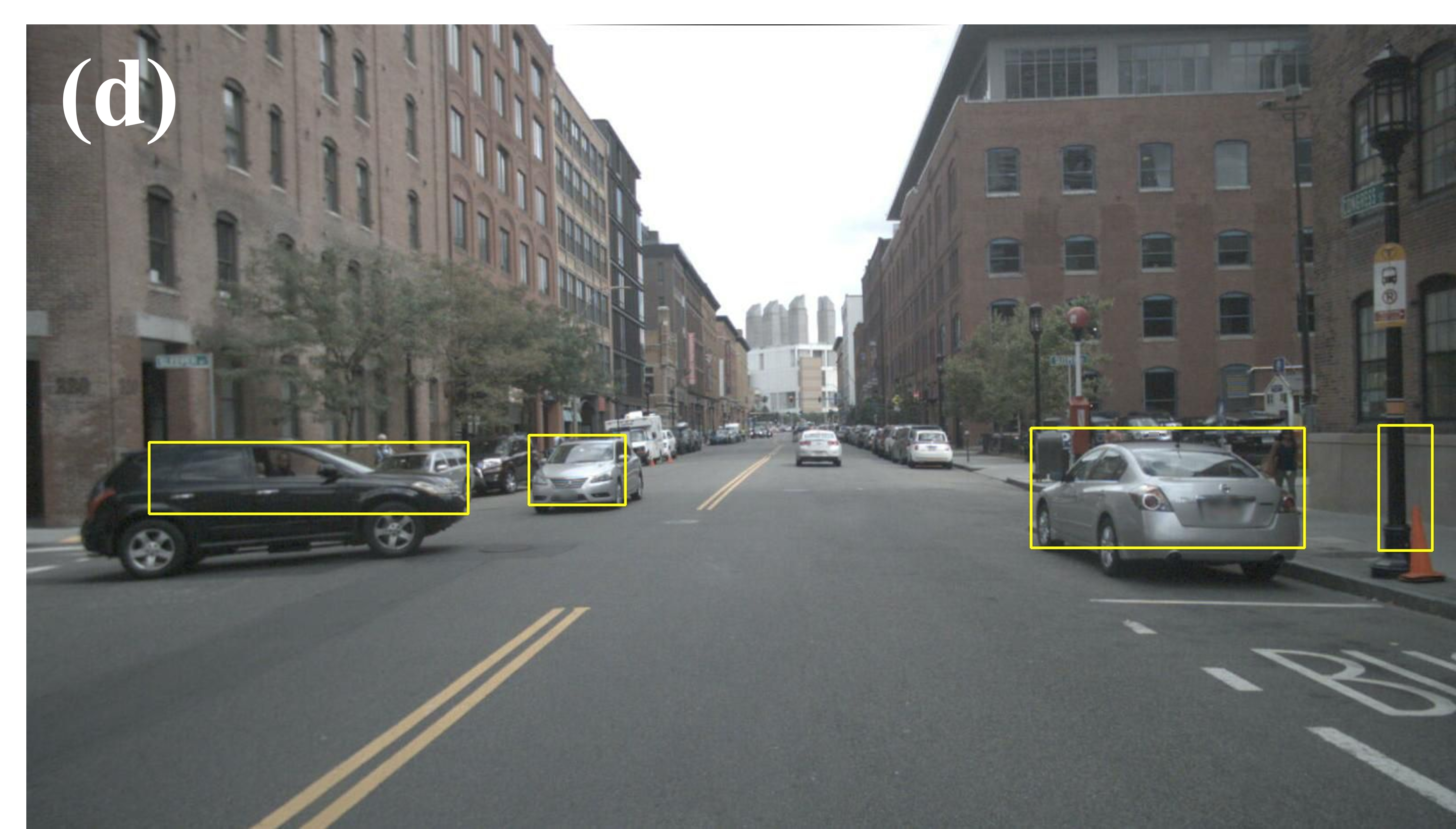}
    \caption{We implement a sequence of image processing techniques to find the safety-critical objects from a given depth map. The images from left to right depict (a) the given depth map, (b) the inverse of the depth map, (c) foreground removal on the inverse, and (d) the bounding boxes rendered on the original image. The first three images have been normalized to the intensity range $[0, 255]$ for visualization.}
    \label{fig:depth_map_processing}
\end{figure*}

To iterate, we attempt to indicate relevant collision risks from object detection results during run time. 
In the following, Sec.~\ref{subsec:object_proposal} introduces how to attain 2.5D objects from a depth map and validates that comparing them with 3D object detection predictions constitutes a valid proxy to ground-truth evaluation. Later, Sec.~\ref{subsec:risk_inference} presents the construction and optimization of the FIS that associates the comparison measures to the output risk level.

\subsection{Depth-based object retrieval and consistency check}
\label{subsec:object_proposal}

As introduced, observing its potential to hint at the whereabouts of objects, we employ a depth estimator that provides absolute depth maps~\cite{bhat2023zoedepth}. Based on the depth maps, a typical object retrieval process consists of two main steps: (i) Apply the Canny edge detecting algorithm to mark the pixels that have large gradient magnitudes~\cite{edge_detecting}; (ii) Find the objects by linking adjacent highlighted pixels into contours~\cite{contour_finding}.

\subsubsection{Preprocessing techniques}
In our work, we exploit the depth information in driving contexts and implement the following two techniques before applying the aforementioned typical process to facilitate better object retrieving results. First, we take the inverse of the raw depth map so as to obtain stronger gradients for pixels closer to the ego vehicle. To explain, with~$I_1(x, y)$ and~$I_2(x, y)$ being two intensity images linearly normalized from the raw depth map~$D(x,y)$ and its inverse~$\frac{1}{D(x,y)}$ respectively, their gradient magnitudes can be written with:
\begin{align}
    & || \nabla I_1(x, y) || \propto || \nabla D(x, y) || \\
    & || \nabla I_2(x, y) || \propto \frac{|| \nabla D(x, y) ||}{D^2(x, y)}, 
\end{align}
where $(x, y)$ denotes a pixel of the image of size $W \times H$. In other words, using the inverse depth map, the same depth value change $|| \nabla D(x, y) ||$ will be scaled into a much smaller gradient magnitude when it is distant from the ego vehicle. This naturally allows the Canny edge detection algorithm to focus on the closer field. The property does not exist with the raw depth map. Fig.~\ref{fig:depth_map_processing}(a) and Fig.~\ref{fig:depth_map_processing}(b) visualizes the comparison. 

One may notice that the foreground close to the ego vehicle is also highlighted after taking the inverse. Fortunately, as an AV always keeps a similar driving perspective, we can remove it by computing the average of all inverse depth maps in the training data and subtracting the average from each map. Together with the first technique, this results in a strong emphasis on the close objects, as shown in Fig.~\ref{fig:depth_map_processing}(c). Lastly, following the typical process (i.e., Canny edge detection and contour linking), we can effectively retrieve the most safety-critical objects in a driving scene. Fig.~\ref{fig:depth_map_processing}(d) shows an example with fitted bounding boxes.

\subsubsection{Measuring alignment with object detector predictions}

Using the retrieved safety-critical objects, we now compare them with the predictions of the underlying object detector. In particular, we focus on the localization attributes and measure how \emph{spatially aligned} the object detector's predictions are with the safety-critical objects.

To start, we represent the safety-critical objects with 2.5D information from the retrieved bounding boxes, e.g., Fig.~\ref{fig:depth_map_processing}(d), and the original depth map, e.g., Fig.~\ref{fig:depth_map_processing}(a). More formally, we denote the set of $M$ safety-critical objects in one frame as $\Tilde{\mathbf{S}} \defeq \{ \hat{\mathbf{S}}_m \, | \, m = 1, \dots, M \}$, where each $\hat{\mathbf{S}} \defeq (x_\mathbf{S}, y_\mathbf{S}, w_\mathbf{S}, h_\mathbf{S}, d_\mathbf{S})$, with $(x_\mathbf{S}, y_\mathbf{S})$ being the center of the bounding box, $w_\mathbf{S}$ the width in the x-axis, $h_\mathbf{S}$ the height in the y-axis, and $d_\mathbf{S}$ the smallest depth value in the area of the bounding box.

Correspondingly, we represent the set of $N$ predicted objects in the frame as $\Tilde{\mathbf{P}} \defeq \{ \hat{\mathbf{P}}_n \, | \, n = 1, \dots, N \}$, where each $\hat{\mathbf{P}} \defeq (x_\mathbf{P}, y_\mathbf{P}, w_\mathbf{P}, h_\mathbf{P}, d_\mathbf{P})$. In practice, a 2D bounding box~$\mathbf{P}$ without the depth attribute can be obtained from a 3D counterpart via perspective projection, and $d_\mathbf{P}$ can be concretized as the closest distance from the 3D bounding box to the AV's position.

\begin{wrapfigure}[8]{r}{0.48\linewidth}
    \centering
    \vspace{-4mm}
    \includegraphics[width=\linewidth]{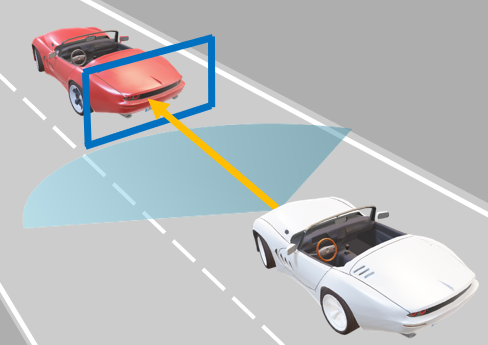}
    \label{fig:safety_factors}
\end{wrapfigure}
Now, to measure the spatial alignment between the two sets~$\Tilde{\mathbf{P}}$ and~$\Tilde{\mathbf{S}}$, we consider two factors, namely the overlap of bounding boxes in the image plane and the depth discrepancy in the driving direction. Visualized in the side figure, these two factors have been shown to be highly relevant to safety in a recent analysis~\cite{mori2024safety}. In practice, we first run the standard Hungarian algorithm to find the closest pairs in terms of their center distances and, for each pair, compute the commonly used Intersection-over-Union (IoU) as well as a simple relative depth discrepancy (RDD) measure:
\begin{align}
    & \mathsf{RDD}( d_\mathbf{P}, d_\mathbf{S} ) \defeq \min ( 1, \frac{ d_\mathbf{P} - d_\mathbf{S} }{d_\mathbf{S}} ) \in [-1, 1].
    \label{eq:rdd}
\end{align}
We additionally apply matching thresholds on the two measures, $\mathsf{IoU} > \alpha$ and $|\mathsf{RDD}| < \beta$, so as to avoid actually distant pairs.
For any unmatched safety-critical objects, the worst scores are assigned, i.e.,~$\mathsf{IoU}=0$ and~$\mathsf{RDD}=1$. Lastly, we aggregate the information in one frame by taking the average of the two measures for all safety-critical objects, denoted as~$\widetilde{\mathsf{IoU}}$ and~$\widetilde{\mathsf{RDD}}$. 

\subsubsection{Interim validation}

To show the computed measures,~$\widetilde{\mathsf{IoU}}$ and~$\widetilde{\mathsf{RDD}}$, are potential for risk inference at the next stage, we now replace~$\Tilde{\mathbf{S}}$ with ground-truth annotations~$\Tilde{\mathbf{G}}$ in the above alignment measuring process and computes the Pearson correlation coefficient between the measures obtained with the safety-critical objects~$\Tilde{\mathbf{S}}$ and those with the ground truths~$\Tilde{\mathbf{G}}$.

For implementation, we use the nuScenes~\cite{caesar2020nuscenes} dataset, the state-of-the-art PGD detector~\cite{wang2021probabilistic}, as well as ZoeDepth estimator~\cite{bhat2023zoedepth}. As the most prominent collision risks stem from nearby objects, we focus on those situated within 20 meters of the ego vehicle. 

\begin{table}[]
    \centering
    \caption{The positive correlation between ``measuring the alignment of the object detector's predictions~$\Tilde{\mathbf{P}}$ and depth-retrieved objects~$\Tilde{\mathbf{S}}$" and ``evaluating the predictions~$\Tilde{\mathbf{P}}$ with ground truths~$\Tilde{\mathbf{G}}$." The correlation increases as the matching thresholds are relaxed in the second row. ($\alpha$, $\beta$) denotes the matching thresholds for $\mathsf{IoU}$ and $\mathsf{RDD}$, respectively.}
    \label{tab:correlation}
    \begin{tabular}{|c|c|c|}
    \hline
    ($\alpha$, $\beta$) & $\widetilde{\mathsf{IoU}}$  & $\widetilde{\mathsf{RDD}}$  \\ \hline
    (0.5, 0.1)          & 0.46 & 0.57 \\ \hline
    (0.3, 0.2)          & 0.65 & 0.72 \\ \hline
    \end{tabular}
\end{table}

Tab.~\ref{tab:correlation} presents the results, from which one can confirm the correlation of interest. Notably, the correlation increases as the alignment matching thresholds are relaxed. This indicates that the depth-based pipeline, while capable of retrieving the objects, may not be precise in terms of the localization coordinates. Thus, for the risk inference in the following, we adopt fuzzy logic, which is known for its strength in handling non-linear functions with imprecision tolerance on the input/output variables.

\subsection{Fuzzy logic-based risk inference}
\label{subsec:risk_inference}

We first introduce the desired risk quantifier, USC, and then illustrate the approaches to construct an optimal FIS for risk estimation.

\subsubsection{A risk-correlating metric as the output target}

Technically, one can use any reasonable risk quantifier as the target. For instance, Wang~\textit{et al.} manually assigns discrete risk levels to driving scenes based on the number of nearby traffic agents~\cite{wang2017collision}. Alternatively, Feth~\textit{et al.} calculates time-to-collision to other vehicles as a proxy~\cite{feth2018dynamic}. In our work, we make use of a safety-oriented object detection metric that has been shown to be strongly correlated with actual AV collision rates in simulations~\cite{liao2024usc}. 

To briefly introduce, the USC metric reflects how well the object detector's predictions \emph{spatially cover} the ground-truth objects when seen from the ego vehicle. Using the representation scheme in Sec.~\ref{subsec:object_proposal}, it is computed for a prediction~$\hat{\mathbf{P}}$ and a ground truth~$\hat{\mathbf{G}}$ as:
\begin{align}
    & \mathsf{USC}(\hat{\mathbf{P}}, \hat{\mathbf{G}}) \defeq \mathsf{IoG}(\mathbf{P}, \mathbf{G}) \times \mathsf{DR}(d_\mathbf{P}, d_\mathbf{G}), \\
    & \text{where } \mathsf{IoG}(\mathbf{P}, \mathbf{G}) \defeq \mathsf{Area}(\mathbf{P} \cap \mathbf{G}) / \mathsf{Area}(\mathbf{G}) \\
    & \text{and } \mathsf{DR}(d_\mathbf{P}, d_\mathbf{G}) \defeq \min (1, d_\mathbf{G} / d_\mathbf{P}).
\end{align}
For an aggregated evaluation on an image frame, we similarly find the average of the USC scores for all ground-truth objects, resulting in $\widetilde{\mathsf{USC}}$. As USC is a scoring function for safety characterization (i.e., the higher, the better), we use the deficit from the full mark, i.e., $1 - \widetilde{\mathsf{USC}}$, as the regression target for risk indication.

\subsubsection{Constructing the FIS}

\begin{figure}
    \centering
    \includegraphics[width=0.98\linewidth]{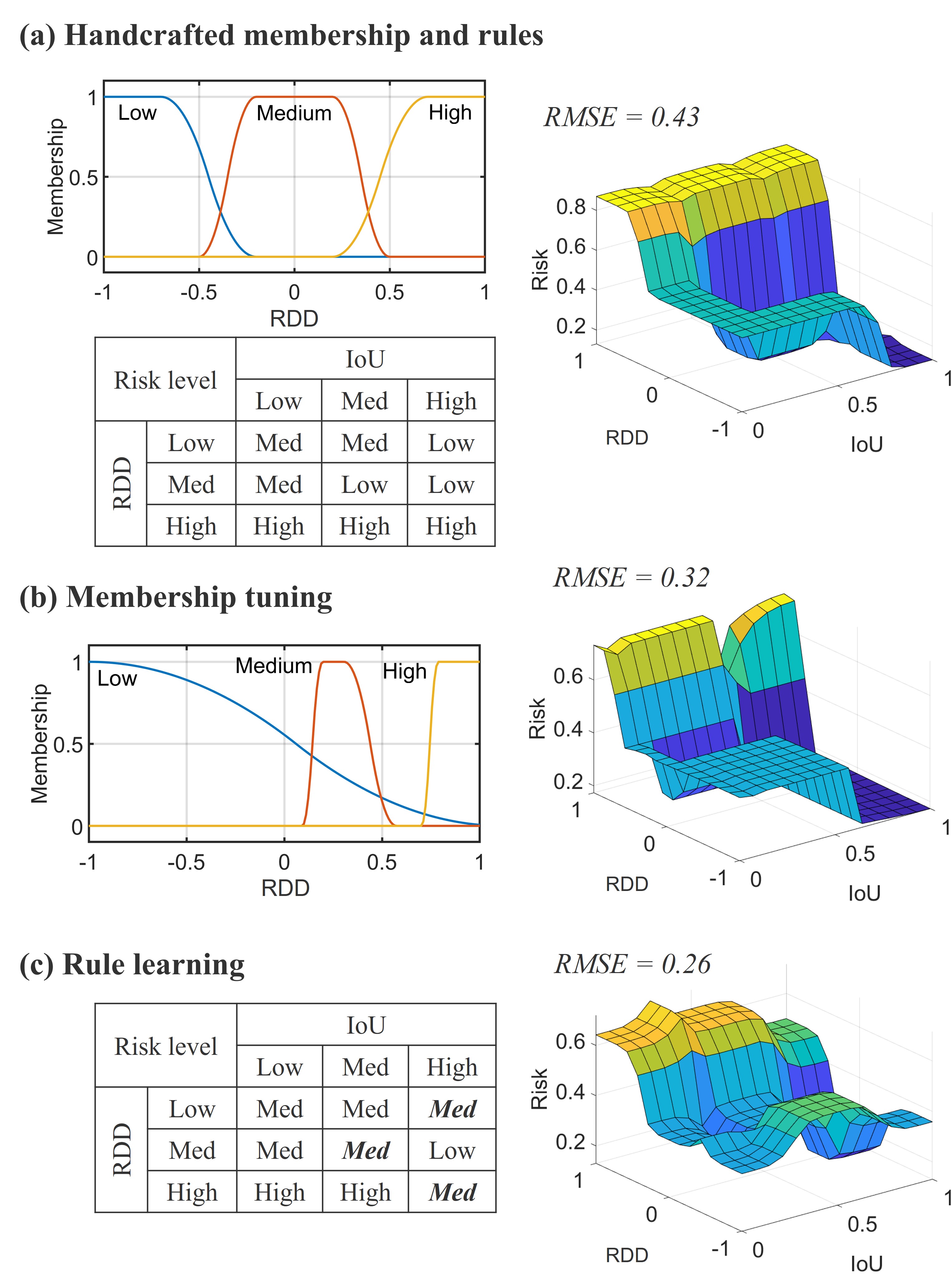}
    \caption{Construction and output surfaces of the FIS via three different approaches, namely (a) knowledge-based crafting, (b) data-driven membership tuning, and (c) data-driven rule learning. For simplicity, the tilde above the variables $\mathsf{RDD}$ and $\mathsf{IoU}$ for denoting frame-wise values are omitted.}
    \label{fig:fis}
\end{figure}

Having formulated the targeted risk quantifier, we describe how a FIS can be constructed to map the alignment measures,~$\widetilde{\mathsf{IoU}}$ and~$\widetilde{\mathsf{RDD}}$, to the target,~$1-\widetilde{\mathsf{USC}}$. 

Essentially, we make use of MATLAB's Mamdani FIS toolbox~\cite{matlab_fis,mamdani1975experiment} and implement three different construction approaches. The first one is based on pure knowledge, specifying (i) high-level fuzzy rules and (ii) membership functions that map real values to linguistic terms for the input/output variables. The second one extends on the first one by having the membership functions optimized with a set of training data via local pattern search. Contrarily, the third one uses the training data to learn a rule set based on the crafted membership functions via global particle swarm optimization. More details about data preparation will be given in Sec.~\ref{sec:validation}.

Fig.~\ref{fig:fis}(a) provides the membership functions (MFs) for~$\widetilde{\mathsf{RDD}}$ and the handcrafted rules based on different combinations of the input variables. To explain, low $\widetilde{\mathsf{RDD}}$'s mean that the predictions are much closer to the ego vehicle than the ground truths, and vice versa. Therefore, by intuition, we assign high risks to high~$\widetilde{\mathsf{RDD}}$ cases. As for medium and low cases, we assign medium or low risks depending on the IoU values. Due to the space limit, MFs for~$\widetilde{\mathsf{IoU}}$ and~$\widetilde{\mathsf{USC}}$ are not shown, but they take the same shapes on the range $[0, 1]$. The resulting output surface of the FIS is displayed aside.

Now, for tuning the MFs, we run the local pattern search algorithm for 1000 iterations using 5-fold cross-validation, as introduced in~\cite{matlab_tuning}. It can be seen from Fig.~\ref{fig:fis}(b) that the ``high" and ``medium" MFs are slightly squeezed to the right, and the ``low'' MF is extended. Correspondingly, the output surface is more sharply decreasing at higher values of~$\widetilde{\mathsf{RDD}}$ and has a larger plateau at the low-$\widetilde{\mathsf{RDD}}$ end. To interpret, the decreasing part is likely because~$1-\widetilde{\mathsf{USC}}$, based on its formulation, tends to tolerate errors and does not often give high values. On the other hand, the larger plateau possibly results from still certain risks when~$\widetilde{\mathsf{RDD}}$ falls around $[-0.5, 0.1]$ (i.e., the left end of the ``medium" function before and after the tuning), as we have linked ``low~$\widetilde{\mathsf{RDD}}$ with medium risk" and ``medium~$\widetilde{\mathsf{RDD}}$ with low risk" when~$\widetilde{\mathsf{IoU}}$ is medium.

The above two tendencies can be observed more apparently from the result of rule learning, for which we execute a global particle swarm optimization algorithm for 20 iterations, each having 100 evaluations~\cite{matlab_tuning}. As Fig.~\ref{fig:fis}(c) shows, the optimizer directly tunes up the risk estimates for two cases when~$\widetilde{\mathsf{RDD}}$ is low and medium. It also decreases the risk estimates in the high-$\widetilde{\mathsf{RDD}}$ high-$\widetilde{\mathsf{IoU}}$ case. To explain, when the object detector recognizes an object in a certain direction correctly (i.e., the high $\widetilde{\mathsf{IoU}}$), even though the depth estimate is far off (i.e., the high $\widetilde{\mathsf{RDD}}$), the AV would drive away from the direction of the object, resulting in smaller collision risks.

Lastly, we conduct evaluations using separate testing data and interestingly find that the FIS built via rule learning can regress the targeted USC metric the best in terms of Root Mean Squared Error (RMSE). One can further improve the results with techniques such as hyperparameter tuning or joint training of the membership functions and the rules. We leave it as future work and will now evaluate the effectiveness of the overall monitoring framework.

\section{Validation and Demonstration}
\label{sec:validation}

\begin{figure*}[]
    \centering
    \includegraphics[width=0.98\linewidth]{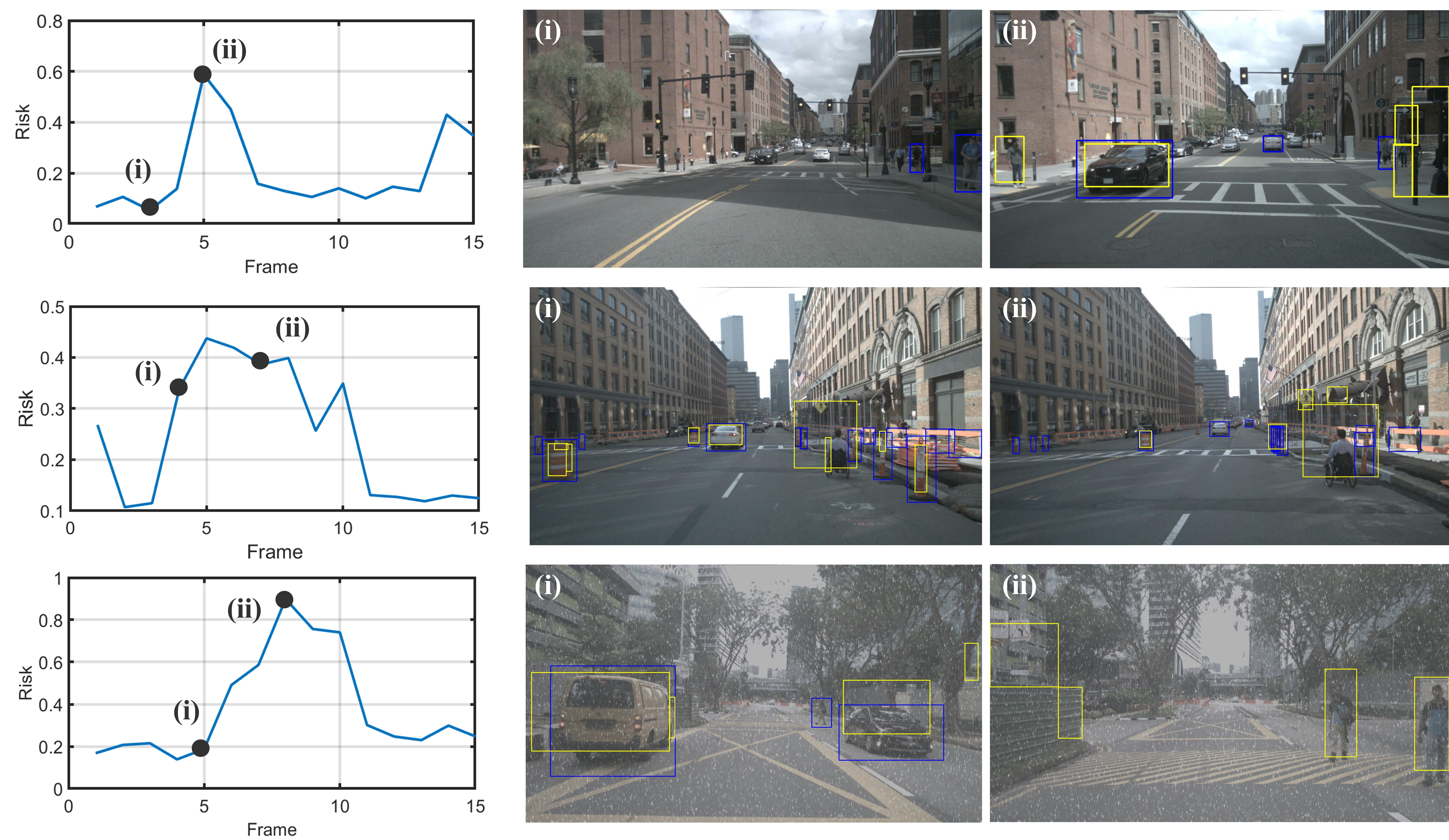}
    \caption{Qualitative results of our monitor applied to the nuScenes dataset~\cite{caesar2020nuscenes}. Blue boxes are from the 3D object detector~\cite{wang2021probabilistic}, and yellow boxes are from the depth-based pipeline. Our monitor successfully translates object detection alignment measures into a useful collision risk indicator.}
    \label{fig:examples}
\end{figure*}

In this section, we first conduct an open-loop validation with image sequences and then demonstrate our monitor in a closed-loop simulation using the globally optimized variant from the above. 

\subsection{Object detector monitoring}
\label{subsec:model}
We continue to use the nuScenes dataset~\cite{caesar2020nuscenes}, the PDG object detector~\cite{wang2021probabilistic}, and the ZoeDepth estimator~\cite{bhat2023zoedepth} for validation. Specifically, we manually selected 110 scenes that involve dense traffic. In addition, to comprehensively validate our monitor, the 110 scenes are diversified into six different conditions, as shown in Tab.~\ref{tab:analysis}. The rare-object subset is built from the annotations in a corner case dataset, CODA~\cite{li2022coda}, and the image perturbation algorithms are employed from a comprehensive benchmark~\cite{dong2023benchmarking}. For FIS testing mentioned in Sec.~\ref{subsec:risk_inference}, four scenes are randomly picked from each condition type; The remaining are the training data.
\begin{table}[]
    \centering
     \caption{Recalls of ground truths~$\Tilde{\mathbf{G}}$ by depth-based objects~$\Tilde{\mathbf{S}}$ and object detector predictions~$\Tilde{\mathbf{P}}$ in various settings. ($\alpha$, $\beta$) denotes the matching thresholds for $\mathsf{IoU}$ and $\mathsf{RDD}$, respectively.}
    \begin{tabular}{|c|c|c|c|c|}
    \hline
    & \begin{tabular}[c]{@{}c@{}}Nominal \\ (0.5, 0.1)\end{tabular} & \begin{tabular}[c]{@{}c@{}}Nominal \\ (0.3, 0.2)\end{tabular} & \begin{tabular}[c]{@{}c@{}}Noise\\ (0.3, 0.2)\end{tabular} & \begin{tabular}[c]{@{}c@{}}Snow\\ (0.3, 0.2)\end{tabular} \\ \hline
    $\Tilde{\mathbf{S}}$ & 0.23                                                          & 0.37 \textit{(+ 61\%)}                                                 & 0.24                                                       & 0.28                                                      \\ \hline
    $\Tilde{\mathbf{P}}$ & 0.41                                                          & 0.48 \textit{(+17\%)}                                                  & 0.21                                                       & 0.23                                                      \\ \hline
    \end{tabular}
    \label{tab:imprecision}
\end{table}

\begin{table}[]
\caption{Detailed analysis on the depth-based object retrieval pipeline, given the object detector's predictions. For each condition, we list the number of scenes, the number of ground truths (GTs), the ratio of approved true positives (TPs), and the ratio of identified false negatives (FPs).}
\label{tab:analysis}
\begin{tabular}{|c|c|c|c|c|}
\hline
Condition      & Scenes & GTs  & Approved TPs & Identified FNs \\ \hline
Nominal        & 25     & 2459 & 881 / 1180   & 382 / 1279     \\ \hline
Rare objects   & 25     & 2129 & 650 / 958    & 521 / 1253     \\ \hline
Gaussian noise & 15     & 1749 & 212 / 368    & 651 / 1381     \\ \hline
Rain           & 15     & 1536 & 402 / 614    & 376 / 922      \\ \hline
Snow           & 15     & 1572 & 129 / 361    & 423 /1211      \\ \hline
Night          & 15     & 1290 & 148 / 795    & 149 / 495      \\ \hline
\end{tabular}
\end{table}

We first evaluate the objects retrieved from the depth-based pipeline~$\Tilde{\mathbf{S}}$ against ground truths. As a benchmark, we take~$\Tilde{\mathbf{P}}$ from the ordinary 3D object detector. It can be seen in Tab.~\ref{tab:imprecision} that in nominal cases, the 3D object detector performs quite better than the depth-based pipeline, but the margin decreases once we allow for the more imprecise predictions. Notably, the depth-based pipeline appears to be more robust than the object detector in the case of sensor noises and snowy conditions, corresponding to recent findings showing the generalization and robustness potential of depth estimators~\cite{xian2024towards}.

Subsequently, Tab.~\ref{tab:analysis} provides a detailed analysis on the depth-based object retrieval pipeline over the object detector's predictions. For each scene condition, we first evaluate the object detector against ground truths and mark the true positives (TPs) as well as false negatives (FNs). Then, we run the depth-based object retrieval pipeline and check how many of the TPs and FNs are retrieved, respectively. Similar to Tab.~\ref{tab:imprecision}, the object detector does not perform well in noisy and snowy conditions, while the depth-based pipeline can recover almost half of the objects. On the contrary, the depth-based pipeline is weaker in night scenes, potentially due to the low illumination, which gives fewer hints for depth estimation.

Lastly, we examine qualitatively the risk estimating output. In Fig.~\ref{fig:examples}, the top row contains a nominal driving scene, where in frame (i), there are multiple detection misses, but our monitor does not raise the risk level due to their distance. By contrast, in frame (ii), several safety-critical pedestrians are not detected, hence the high risk. For the middle scene, a person in a wheelchair continues to be missed by the ordinary object detector. The depth-based pipeline can detect the person and trigger a high-risk output. Finally, the bottom row shows a simulated snowy scene, where two jaywalkers are missed by the 3D object detector yet recovered by the depth-based pipeline.

\subsection{AV safeguarding}
\label{subsec:system}
\begin{figure}
    \centering
    \includegraphics[width=0.98\linewidth]{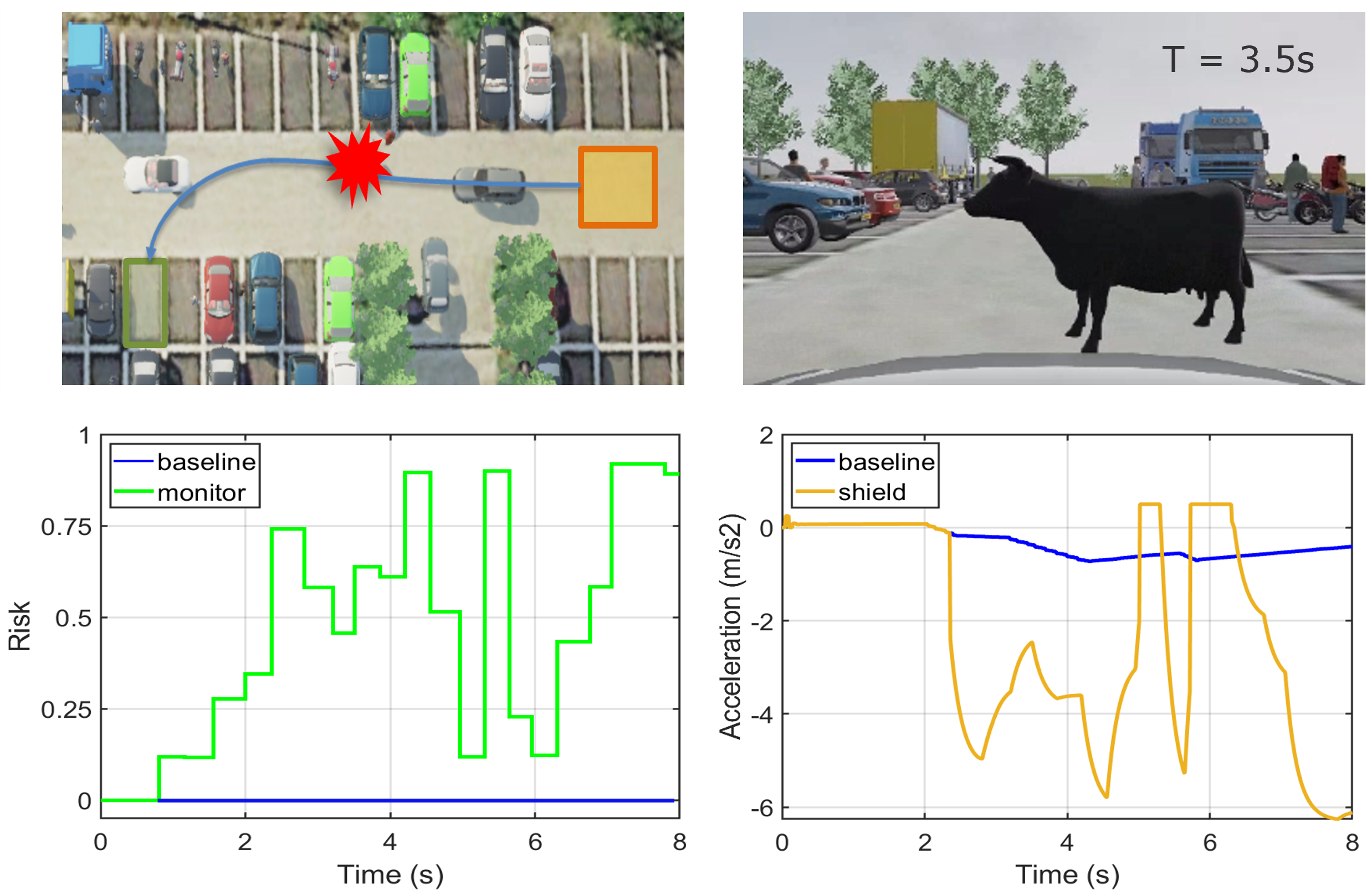}
    \caption{In a simulated scenario with an unseen cow~\textit{(top)}, our monitor successfully raises the risk level and triggers a controller shield to decelerate the AV \textit{(bottom; green)}. On the contrary, the baseline AV without the monitor and the shield would continue to drive and hit the cow \textit{(bottom; blue)}.}
    \label{fig:demo}
\end{figure}

Finally, we demonstrate our monitor's capability to safeguard an AV in unusual situations. To briefly illustrate, a baseline AV has been designed for an automated valet parking (AVP) use case in the Prescan simulator~\cite{bensalem2023continuous}. In each mission, the AV has to drive from a hand-off zone to a parking slot while navigating through mixed traffic of other vehicles and pedestrians. In addition, to mimic real-world operating conditions, different effects such as low illumination, weather changes, and rare objects are added to the simulations. For instance, Fig.~\ref{fig:demo} shows a scenario with a cow that has not been seen by the baseline object detector, likely ending up as a false negative. 

Now, to protect the AV against such challenging scenarios, we implement our perception monitor alongside a controller shield on top of the baseline AV stack. Whenever our monitor raises the risk level beyond a threshold (empirically set as 0.5), the controller shield starts to decelerate the AV. Fig.~\ref{fig:demo} depicts that our monitor successfully triggers the shield to decelerate the AV, which would have crashed into the cow. In general, we observe a successful collision avoidance rate beyond 80\%. For a more lively demonstration with another scenario, readers can check a full use case video~(where our run-time monitor is showcased from 9:16 minute)~\cite{liao2023demo}.

\section{Conclusion}

In this work, by observing and validating that the alignment measures between an ordinary object detector's predictions and the objects retrieved from a depth-based pipeline can be correlated to the errors of the object detector against ground truths, we implemented a monitoring framework to estimate ego vehicle's collision risks based on the alignment measures. Specifically, we used a risk-correlating metric to optimize a fuzzy inference system for the run-time collision risk estimation. As a result, we are able to infer a relevant collision risk level using low-cost monocular camera images only. Experiments with the nuScenes dataset and a closed-loop simulator demonstrated the effectiveness of our approach. 

Our work opens several interesting directions to explore. For instance, one may use additional signals such as scene flow estimates to find objects that move fast towards the ego vehicle. In addition, intuitive rules regarding object classification results can also be considered, e.g., a car object should normally occupy $x$ pixels at distance $y$. Lastly, we plan to investigate how the monitoring results can be extended with active learning techniques to improve the perception results and, hence, the AV's performance.

\balance

\bibliographystyle{IEEEtran}
\bibliography{ref}

\begin{thebibliography}{10}
\providecommand{\url}[1]{#1}
\csname url@samestyle\endcsname
\providecommand{\newblock}{\relax}
\providecommand{\bibinfo}[2]{#2}
\providecommand{\BIBentrySTDinterwordspacing}{\spaceskip=0pt\relax}
\providecommand{\BIBentryALTinterwordstretchfactor}{4}
\providecommand{\BIBentryALTinterwordspacing}{\spaceskip=\fontdimen2\font plus
\BIBentryALTinterwordstretchfactor\fontdimen3\font minus
  \fontdimen4\font\relax}
\providecommand{\BIBforeignlanguage}[2]{{%
\expandafter\ifx\csname l@#1\endcsname\relax
\typeout{** WARNING: IEEEtran.bst: No hyphenation pattern has been}%
\typeout{** loaded for the language `#1'. Using the pattern for}%
\typeout{** the default language instead.}%
\else
\language=\csname l@#1\endcsname
\fi
#2}}
\providecommand{\BIBdecl}{\relax}
\BIBdecl

\bibitem{vanzura2024autonomous}
M.~Vanžura, M.~Bureš, and K.~Metivier, ``Autonomous vehicle crashes,''
  \url{https://www.avcrashes.net/}, 2024.

\bibitem{aiact}
``{Artificial Intelligence Act},''
  \url{https://eur-lex.europa.eu/legal-content/EN/TXT/?uri=OJ:L_202401689},
  2024.

\bibitem{iso21448}
``{ISO} 21448:2022 {R}oad vehicles - {S}afety of the intended functionality,''
  \url{https://www.iso.org/standard/77490.html}, 2022.

\bibitem{chia2022risk}
W.~M.~D. Chia, S.~L. Keoh, C.~Goh, and C.~Johnson, ``Risk assessment
  methodologies for autonomous driving: A survey,'' \emph{T-ITS}, 2022.

\bibitem{chen2021monitoring}
Y.~Chen, C.-H. Cheng, J.~Yan, and R.~Yan, ``Monitoring object detection
  abnormalities via data-label and post-algorithm abstractions,'' in
  \emph{IROS}, 2021.

\bibitem{balakrishnan2021percemon}
A.~Balakrishnan, J.~Deshmukh, B.~Hoxha, T.~Yamaguchi, and G.~Fainekos,
  ``{PerceMon}: {O}nline monitoring for perception systems,'' in \emph{RV},
  2021.

\bibitem{liao2024usc}
B.~H.-C. Liao, C.-H. Cheng, H.~Esen, and A.~Knoll, ``{USC}: {U}ncompromising
  spatial constraints for safety-oriented 3{D} object detectors in autonomous
  driving,'' in \emph{ITSC}, 2024.

\bibitem{bhat2023zoedepth}
S.~F. Bhat, R.~Birkl, D.~Wofk, P.~Wonka, and M.~Müller, ``{ZoeDepth}:
  {Z}ero-shot transfer by combining relative and metric depth,'' 2023.

\bibitem{zadeh1988fuzzy}
L.~Zadeh, ``Fuzzy logic,'' \emph{Computer}, 1988.

\bibitem{wang2021probabilistic}
T.~Wang, X.~Zhu, J.~Pang, and D.~Lin, ``Probabilistic and geometric depth:
  {D}etecting objects in perspective,'' in \emph{CoRL}, 2021.

\bibitem{caesar2020nuscenes}
H.~Caesar, V.~Bankiti, A.~H. Lang, S.~Vora, V.~E. Liong, Q.~Xu, A.~Krishnan,
  Y.~Pan, G.~Baldan, and O.~Beijbom, ``nu{S}cenes: {A} multimodal dataset for
  autonomous driving,'' in \emph{CVPR}, 2020.

\bibitem{zheng2018novel}
X.~Zheng, B.~Huang, D.~Ni, and Q.~Xu, ``A novel intelligent vehicle risk
  assessment method combined with multi-sensor fusion in dense traffic
  environment,'' \emph{JICV}, 2018.

\bibitem{khonji2020risk}
M.~Khonji, J.~Dias, R.~Alyassi, F.~Almaskari, and L.~Seneviratne, ``A
  risk-aware architecture for autonomous vehicle operation under uncertainty,''
  in \emph{SSRR}, 2020.

\bibitem{wang2017collision}
Y.~Wang and J.~Kato, ``Collision risk rating of traffic scene from dashboard
  cameras,'' in \emph{DICTA}, 2017.

\bibitem{feth2018dynamic}
P.~Feth, M.~N. Akram, R.~Schuster, and O.~Wasenm{\"u}ller, ``Dynamic risk
  assessment for vehicles of higher automation levels by deep learning,'' in
  \emph{SAFECOMP Workshops}, 2018.

\bibitem{grimmett2016introspective}
H.~Grimmett, R.~Triebel, R.~Paul, and I.~Posner, ``Introspective classification
  for robot perception,'' \emph{The International Journal of Robotics
  Research}, 2016.

\bibitem{hendrycks17baseline}
D.~Hendrycks and K.~Gimpel, ``A baseline for detecting misclassified and
  out-of-distribution examples in neural networks,'' in \emph{ICLR}, 2017.

\bibitem{gal2016dropout}
Y.~Gal and Z.~Ghahramani, ``Dropout as a bayesian approximation: {R}epresenting
  model uncertainty in deep learning,'' in \emph{ICML}, 2016.

\bibitem{lakshminarayanan2017simple}
B.~Lakshminarayanan, A.~Pritzel, and C.~Blundell, ``Simple and scalable
  predictive uncertainty estimation using deep ensembles,'' in \emph{NeurIPS},
  2017.

\bibitem{le2018uncertainty}
M.~T. Le, F.~Diehl, T.~Brunner, and A.~Knoll, ``Uncertainty estimation for deep
  neural object detectors in safety-critical applications,'' in \emph{ITSC},
  2018.

\bibitem{nallapareddy2023evcenternet}
M.~R. Nallapareddy, K.~Sirohi, P.~L.~J. Drews, W.~Burgard, C.-H. Cheng, and
  A.~Valada, ``{EvCenterNet}: {U}ncertainty estimation for object detection
  using evidential learning,'' in \emph{IROS}, 2023.

\bibitem{du2022vos}
X.~Du, Z.~Wang, M.~Cai, and Y.~Li, ``{VOS}: {L}earning what you don’t know by
  virtual outlier synthesis,'' in \emph{ICLR}, 2022.

\bibitem{wu2024bam}
C.~Wu, W.~He, C.-H. Cheng, X.~Huang, and S.~Bensalem, ``{BAM}: {B}ox
  abstraction monitors for real-time {OoD} detection in object detection,'' in
  \emph{IROS}, 2024.

\bibitem{rahman2021online}
Q.~M. Rahman, N.~Sünderhauf, and F.~Dayoub, ``Online monitoring of object
  detection performance during deployment,'' in \emph{IROS}, 2021.

\bibitem{otani2022optimal}
M.~Otani, R.~Togashi, Y.~Nakashima, E.~Rahtu, J.~Heikkilä, and S.~Satoh,
  ``Optimal correction cost for object detection evaluation,'' in \emph{CVPR},
  2022.

\bibitem{pandharipande2023sensing}
A.~Pandharipande, C.-H. Cheng, J.~Dauwels, S.~Z. Gurbuz, J.~Ibanez-Guzman,
  G.~Li, A.~Piazzoni, P.~Wang, and A.~Santra, ``Sensing and machine learning
  for automotive perception: A review,'' \emph{IEEE Sens. J.}, 2023.

\bibitem{edge_detecting}
``Canny edge detection,''
  \url{https://docs.opencv.org/4.10.0/da/d22/tutorial_py_canny.html?ref=blog.roboflow.com}.

\bibitem{contour_finding}
``Contours,''
  \url{https://docs.opencv.org/4.10.0/d4/d73/tutorial_py_contours_begin.html}.

\bibitem{mori2024safety}
K.~T. Mori and S.~Peters, ``{SHARD: Safety} and human performance analysis for
  requirements in detection,'' \emph{T-IV}, 2024.

\bibitem{matlab_fis}
``Fuzzy inference process,''
  \url{https://de.mathworks.com/help/fuzzy/fuzzy-inference-process.html#FP346}.

\bibitem{mamdani1975experiment}
E.~Mamdani and S.~Assilian, ``An experiment in linguistic synthesis with a
  fuzzy logic controller,'' \emph{Int. J. Man-Mach. Stud.}, 1975.

\bibitem{matlab_tuning}
``Tuning fuzzy inference systems,''
  \url{https://de.mathworks.com/help/fuzzy/tune-fuzzy-inference-systems.html}.

\bibitem{li2022coda}
K.~Li, K.~Chen, H.~Wang, L.~Hong, C.~Ye, J.~Han, Y.~Chen, W.~Zhang, C.~Xu,
  D.-Y. Yeung, X.~Liang, Z.~Li, and H.~Xu, ``Coda: A real-world road corner
  case dataset for object detection in autonomous driving,'' in \emph{ECCV},
  2022.

\bibitem{dong2023benchmarking}
Y.~Dong, C.~Kang, J.~Zhang, Z.~Zhu, Y.~Wang, X.~Yang, H.~Su, X.~Wei, and
  J.~Zhu, ``Benchmarking robustness of {3D} object detection to common
  corruptions in autonomous driving,'' in \emph{CVPR}, 2023.

\bibitem{xian2024towards}
C.~S. . G.~L. Ke~Xian, Zhiguo~Cao, ``Towards robust monocular depth estimation:
  {A} new baseline and benchmark,'' 2024.

\bibitem{bensalem2023continuous}
S.~Bensalem, P.~Katsaros, D.~Ni{\v{c}}kovi{\'{c}}, B.~H.-C. Liao, R.~R.
  Nolasco, M.~A. E.~S. Ahmed, T.~A. Beyene, F.~Cano, A.~Delacourt, H.~Esen,
  A.~Forrai, W.~He, X.~Huang, N.~Kekatos, B.~K{\"o}nighofer, M.~Paulitsch,
  D.~Peled, M.~Ponchant, L.~Sorokin, S.~Tong, and C.~Wu, ``Continuous
  engineering for trustworthy learning-enabled autonomous systems,'' in
  \emph{{AISoLa}}, 2023.

\bibitem{liao2023demo}
B.~H.-C. Liao, ``{S}afety assurance for an automated valet parking system,''
  \url{https://youtu.be/ngqIFae4Qeg?si=f6Fmkqkt76lV-pxz}, 2023.

\end{thebibliography}

\end{document}